\documentclass{article}

\usepackage{PRIMEarxiv}
\usepackage[utf8]{inputenc} 
\usepackage[T1]{fontenc}    
\usepackage{hyperref}       
\usepackage{url}            
\usepackage{booktabs}       
\usepackage{amsfonts}       
\usepackage{nicefrac}       
\usepackage{microtype}      
\usepackage{lipsum}
\usepackage{fancyhdr}       
\usepackage{graphicx}       
\graphicspath{{media/}}     
\usepackage{xspace}
\usepackage{amsmath}
\usepackage{algorithm}  
\usepackage{algorithmicx}  
\usepackage{algpseudocode}
\usepackage{multirow}
\usepackage{subfigure}
\usepackage{fontawesome}
\usepackage{color}
\newcommand{\nickname}{\textit{semi-PD}\xspace}
\newcommand{\hk}[1]{\textcolor{black}{#1}}

\title{\nickname: Towards Efficient LLM Serving via Phase-Wise Disaggregated Computation and Unified Storage
}

\author{
  Ke Hong$^{\text{1,2}}$\\
  \And
  Lufang Chen$^{\text{2}}$\\
  \And 
  Zhong Wang$^{\text{2}}$\\
  \And
  Xiuhong Li$^{\text{2}}$~$^{\text{\faEnvelope}}$ \\
  \And
  Qiuli Mao$^{\text{2}}$\\
  \And 
  Jianping Ma$^{\text{2}}$\\
  \And
  Chao Xiong$^{\text{2}}$\\
  \And 
  Guanyu Wu$^{\text{1,2}}$\\
  \And 
  Buhe Han$^{\text{2}}$\\
  \And
  Guohao Dai$^{\text{2,3}}$~$^{\text{\faEnvelope}}$\\
  \And 
  Yu Wang$^{\text{1}}$~$^{\text{\faEnvelope}}$\\
  \And
  \large{{\faEnvelope~}\texttt{lixiuhong@infini-ai.com, daiguohao@sjtu.edu.cn, yu-wang@tsinghua.edu.cn}}
}
\begin{document}
\maketitle
\def\thefootnote{$^1$}\footnotetext{Tsinghua University}\def\thefootnote{\arabic{footnote}}
\def\thefootnote{$^2$}\footnotetext{Infinigence-AI}\def\thefootnote{\arabic{footnote}}
\def\thefootnote{$^3$}\footnotetext{Shanghai Jiao Tong University}\def\thefootnote{\arabic{footnote}}
\def\thefootnote{\faEnvelope}\footnotetext{Corresponding authors: Yu Wang, Xiuhong Li, Guohao Dai.}\def\thefootnote{\arabic{footnote}}

\begin{abstract}
Existing large language model (LLM) serving systems fall into two categories: 1) a unified system where \textit{prefill} phase and \textit{decode} phase are co-located on the same GPU, sharing the unified computational resource and storage, and 2) a disaggregated system where the two phases are disaggregated to different GPUs. The design of the disaggregated system addresses the latency interference and sophisticated scheduling issues in the unified system but leads to storage challenges including 1) replicated weights for both phases that prevent flexible deployment, 2) KV cache transfer overhead between the two phases, 3) storage imbalance that causes substantial wasted space of the GPU capacity, and 4) suboptimal resource adjustment arising from the difficulties in migrating KV cache. Such storage inefficiency delivers poor serving performance under high request rates. 

In this paper, we identify that the advantage of the disaggregated system lies in the disaggregated computation, \textit{i.e.}, partitioning the computational resource to enable the asynchronous computation of two phases. Figure~\ref{fig:abstrct} illustrates the pros and cons of the different computation and storage patterns. Thus, we propose a novel LLM serving system, \nickname, characterized by disaggregated computation and unified storage. In \nickname, we introduce a computation resource controller to achieve disaggregated computation at the streaming multi-processor (SM) level, and a unified memory manager to manage the asynchronous memory access from both phases. \nickname has a low-overhead resource adjustment mechanism between the two phases. On top of such an adjustment mechanism, we further design a service-level objective (SLO) aware dynamic partitioning algorithm to optimize the SLO attainment. The evaluation shows that compared to state-of-the-art systems,  \nickname maintains lower latency at higher request rates, \hk{reducing the average end-to-end latency per request by 1.27-2.58$\times$ on DeepSeek series models, and serves 1.55-1.72$\times$ more requests adhering to latency constraints on Llama series models.}
\end{abstract}


\section{Introduction}


With the rapid development of large language models, more and more users are beginning to adopt large language model services for content generation, in the domains including chatbots~\cite{brown2020language,adiwardana2020towards,roller2021recipes}, code assistants~\cite{feng2020codebert,svyatkovskiy2020intellicode}, and agent-based applications~\cite{bubeck2023sparks,ouyang2022training,openai2023gpt4}.
In LLM serving, users typically send requests to large model service providers and wait for a response. However, LLM service tends to introduce a considerable latency due to the tremendous computation and memory access requirements~\cite{survey2024}.
Unlike traditional requests, the latency of LLM requests consists of two distinct parts. This is because each LLM request involves two processing phases: the \textit{prefill} phase and the \textit{decode} phase.
The \textit{prefill} phase processes the input prompt and generates the first output token, and the \textit{decode} phase generates the output tokens iteration by iteration in an auto-regressive manner, with one token at each iteration.
Thus, the first part is the Time-to-First-Token (TTFT), which measures the time between the request arrival and the first output token being generated via the \textit{prefill} phase.
The second part is the Time-per-Output-Token (TPOT), which measures the average latency per token in the \textit{decode} phase.
Constraining TTFT and TPOT, as the two components of the service-level objective (SLO), are both essential for user experience.
Thus, adhering to the TTFT and TPOT constraints, how to schedule the \textit{prefill} and \textit{decode} phase of each request to maximize the system throughput is very important.

\begin{figure}
    \centering
    \includegraphics[width=0.75\linewidth]{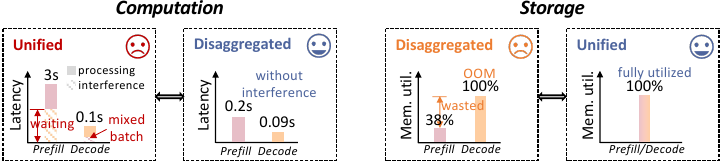}
    \caption{Illustration of the pros and cons of the different computation and storage
patterns. \nickname can have the advantages of both disaggregated computation and unified storage.}
    \label{fig:abstrct}
\end{figure}

\begin{figure*}[t]
    \centering
    \includegraphics[width=0.95\linewidth]{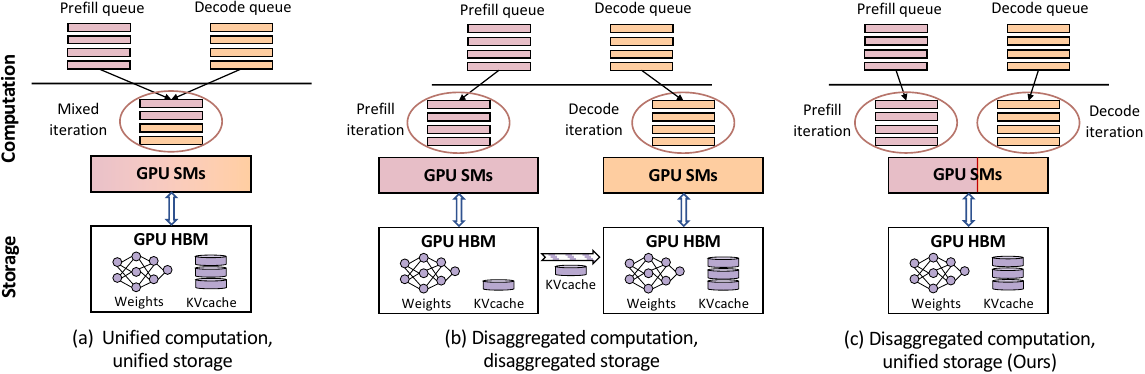}
    \vspace{-0.5em}
    \caption{Comparison among different computation and storage design paradigms for LLM serving.}
    \vspace{-0.5em}
    \label{fig:intro_1}
\end{figure*}

LLM request batching is an effective method for improving system throughput.
First of all, to solve the challenge that the LLM requests have different input prompt lengths and different output generation lengths, Orca~\cite{yu2022orca} proposes the continuous batching technique, scheduling execution and batching requests at the granularity of iterations (an execution of \textit{prefill} phase or \textit{decode} phase) instead of requests.
Based on the iterative batching mechanism, many works focus on how to schedule \textit{prefill} iterations and \textit{decode} iterations.
Researchers first choose to place \textit{prefill} and \textit{decode} iterations within the same GPU as shown in Figure~\ref{fig:intro_1}(a).
They either prioritize the \textit{prefill} iterations (vLLM~\cite{kwon2023vllm}) or prioritize the \textit{decode} iterations (FasterTransformer~\cite{FasterTransformer}).
The former benefits TTFT but at the cost of TPOT, and vice versa. Such a trade-off is called the latency interference between the two phases. 
Therefore, to alleviate the interference of TTFT on TPOT, Sarathi-Serve~\cite{agrawal2023sarathi} and DeepSpeed-FastGen~\cite{holmes2024deepspeedfastgen} propose to chunk the input of \textit{prefill} request into smaller \textit{prefill} segments and then schedule the \textit{prefill} segments together with \textit{decode} requests in one batch.
However, the aforementioned works cannot avoid the interference between TTFT and TPOT, and it is very challenging to meet both TTFT and TPOT constraints in SLO with an optimal system throughput.
To this end, researchers propose the disaggregated design which places the \textit{prefill} iteration and \textit{decode} iteration on different GPUs to eliminate the interference between TTFT and TPOT~\cite{hu2024tetriinfer, patel2023splitwise, zhong2024distserve, oh2024exegpt, qin2024mooncake}.
In the disaggregated system, some GPUs act as the \textit{prefill} instance, and others act as the \textit{decode} instance, as shown in Figure~\ref{fig:intro_1}(b).

\hk{In this paper, we use the term storage to denote the memory capacity on GPUs.} We identify the drawbacks of the disaggregated design caused by its disaggregated storage for the model weights and KV cache.
1)~\textbf{Storage imbalance.} 
The \textit{prefill} instance only generates the part of the KV cache corresponding to the input length. Meanwhile, the \textit{decode} instance needs to accommodate the whole KV cache corresponding to the full sequence length including the output length.
In cases where the output length is larger, the storage imbalance will become more severe. Thus, the storage capacity required by the \textit{decode} instance is significantly larger than that of the \textit{prefill} instance.
2)~\textbf{KV cache transfer.} Since the \textit{prefill} phase and \textit{decode} phase are placed on different GPUs, it is necessary to transfer the KV cache from the \textit{prefill} instance to the \textit{decode} instance.
The transfer overhead becomes significant on low-end GPUs due to communication latency, necessitating careful parallelism strategy designs~\cite{zhong2024distserve}.
Even for the high-end GPUs equipped with NVLink~\cite{nvlink2024}, the transfer time will be calculated into TPOT, making TPOT sub-optimal.
3)~\textbf{Resource adjustment between \textit{prefill} and \textit{decode} instances.} The workload of \textit{prefill} and \textit{decode} requests varies as the serving process goes on. However, the GPU-level disaggregation results in coarse-grained adjustment. Moreover, since the KV cache is mainly located on the \textit{decode} instance, the adjustment needs specific designs to avoid the cost of KV cache transfer, which introduces additional overhead.
4)~\textbf{Weight replica.} The \textit{prefill} instance and \textit{decode} instance both need to accommodate the whole model weight. Disaggregated systems face deployment challenges when the number of available GPUs is limited. 

We comprehensively compare and analyze the unified design and disaggregated design, as shown in Figure~\ref{fig:intro_1}(a) and Figure~\ref{fig:intro_1}(b).
The key insight is that we should adopt the disaggregated computation feature of the disaggregated system and reserve the unified storage feature of the unified system.
Moreover, the system is supposed to be lightweight and support adjusting the resource ratio between \textit{prefill} and \textit{decode} phases with negligible overhead, thereby adaptively matching the workload changes during serving.
In this paper, we propose \nickname, an efficient LLM serving system characterized by phase-wise disaggregated computation and unified storage.
The design of \nickname consists of 1) a computation resource controller to support disaggregated computation and lightweight resource adjustment, with the disaggregated computation of each phase abstracted as a worker, and 2) a unified memory manager to coordinate the memory accesses of model weights and KV cache for both \textit{prefill} and \textit{decode} workers.
Based on that, we orchestrate a low-overhead dynamic resource-adjusting algorithm based on the SLO requirements.
Finally, we implement \nickname and perform a comprehensive evaluation to compare with the state-of-the-art works.

In summary, this paper makes the following contributions:
\begin{itemize}
\item We comprehensively analyze the unified and disaggregated designs and propose a phase-wise disaggregated computation and unified storage design to improve serving performance. We further build a system prototype that fully matches the proposed design.

\item Our system supports adjusting the resource ratio between \textit{prefill} and \textit{decode} workers with a low-overhead worker switching mechanism and SLO-aware switching algorithm.

\item We implement \nickname, and evaluate across various benchmarks. The results demonstrate that \nickname \hk{reduces the request-level latency by 1.27-2.58$\times$}, and serves 1.55-1.72$\times$ more requests, compared to state-of-the-art systems while staying within latency constraints for > 90\% of requests.

\end{itemize}

\section{Background}
\subsection{LLM Processing Phases}

Modern LLMs~\cite{openai2023gpt4, touvron2023llama} utilize transformer-based~\cite{vaswani2017attention} decoder-only~\cite{brown2020gpt3} architecture to predict the next token in an autoregressive manner. The LLM processing can be divided into two phases: \textit{prefill} phase and \textit{decode} phase. Such design benefits from the KV cache that stores the information of previous tokens to avoid repeated computation. During the \textit{prefill} phase, the model processes the input context, generates the initial KV cache for each input token, and derives the first output token. During the \textit{decode} phase, the model produces the next token based on the last generated token and the KV cache. Meanwhile, the newly generated key and value vectors of the current input token are appended to the KV cache. The LLM iteratively generates output tokens until reaches a pre-defined maximum token length or an end-of-sequence (\emph{<eos>}) token. 

\subsection{LLM Serving}

In LLM serving, users interact with the LLM by sending requests (\textit{e.g.}, text inputs) and receiving responses (\textit{e.g.}, generated text). 

\subsubsection{Latency and SLOs.}\label{sec:metrics}
Commonly used latency metrics for LLM serving include TTFT and TPOT.
\begin{itemize}
    \item Time-to-First-Token (TTFT) is the duration between the arrival time and the time when the \textit{prefill} phase finishes of a request.
    \item Time-per-Output-Token (TPOT) is the average duration to generate the subsequent tokens (except for the first one) of a request. Some works~\cite{qin2024mooncake} use Time-between-Tokens (TBT) to measure the latency between successive tokens. In this work, we use the TPOT metric.  
\end{itemize}
TTFT and TPOT capture the request-level latency experienced by users. To ensure the user experience, the agreement between service providers and users determines SLOs (\textit{e.g.}, TTFT $\leq$ 500ms, TPOT $\leq$ 100ms) as the latency constraints. To address the importance of SLOs, many works ~\cite{agrawal2024sarathiserve, zhong2024distserve, wu2024loongserve, qin2024mooncake} consider serving more requests while adhering to the SLOs, which is also the objective of this work. 

\subsubsection{Batching.}\label{sec:batching}

Continuous batching is the basic technique in LLM serving. The differences in the output length prevent the requests in a batch from being finished at the same time. The ended requests are not returned to users immediately, and the incoming requests are forced to wait until the current batch is fully processed. That kind of request-level batching harms both system-level utilization and request-level latency. To address the issue, the technique of continuous batching is introduced in Orca~\cite{yu2022orca}, which batches the requests at the iteration level. With continuous batching, at each iteration, the finished requests exit, and the incoming requests are added to the current batch. The continuous batching technique is widely used in various serving systems, including both unified and disaggregated designs. 

\subsection{Related Works}


Based on the continuous batching technique, different LLM serving designs have been proposed. In this section, we discuss the related works including the phase-wise unified and disaggregated system designs. Moreover, we also list the related works involving request prioritization and throughput-oriented optimization, which are orthogonal to this work. 

\subsubsection{Unified system.} In a unified system, the works with batching and scheduling designs mainly focus on the trade-offs between \textit{prefill} phase and \textit{decode} phase. The default scheduling strategy in vLLM~\cite{kwon2023vllm} prioritizes the \textit{prefill} phase of new requests over the \textit{decode} phase of the running ones to improve the throughput via larger batch size, but at the cost of TPOT degrading drastically. FasterTransformer~\cite{FasterTransformer} prioritizes \textit{decode} iteration, leading to better TPOT at the cost of worse TTFT and system throughput. To mitigate such degradation, Sarathi-Serve~\cite{agrawal2024sarathiserve} and DeepSpeed-FastGen~\cite{holmes2024deepspeedfastgen} propose the SplitFuse method to chunk the \textit{prefill} requests along the sequence dimension and schedule the \textit{decode} requests together with the chunked \textit{prefill} requests within one batch. Besides, SGLang~\cite{zheng2023sglang}, TensorRT-LLM~\cite{tensorrt-llm}, and LMDeploy~\cite{lmdeploy} are all highly optimized unified systems for LLM serving. Nevertheless, all of those implementations fail to eliminate the latency interference.

\subsubsection{Disaggregated system.} The disaggregated design removes the \textit{prefill}-\textit{decode} interference by disaggregating the \textit{prefill} and \textit{decode} phases to different GPUs. 
Based on the disaggregated design, Splitwise~\cite{patel2023splitwise} further utilizes different types of GPU for different phase deployment and maintains a mixed machine pool to realize the resource adjustment between phases. DistServe~\cite{zhong2024distserve} develops algorithms to derive the optimal parallelism separately for \textit{prefill} phase and \textit{decode} phase and reruns the placement algorithm periodically to suit the workload changes. TetriInfer~\cite{hu2024tetriinfer} presents a detailed top-down system to discuss the design against various workloads and use an instance flip mechanism for resource adjustment between the two phases. 
ExeGPT~\cite{oh2024exegpt} proposes different scheduling strategies involving assigning the execution of \textit{prefill} and \textit{decode} phases to different GPUs, and discusses the performance under the latency constraint.
Mooncake~\cite{qin2024mooncake} and MemServe~\cite{hu2024memserve} both unveil the potential of disaggregated systems combined with the prefix caching technique~\cite{zheng2023sglang}.  
LoongServe~\cite{wu2024loongserve} introduces elastic sequence parallelism, where each machine stores a portion of the KV cache, supporting more flexible dynamic scaling up or down.
\hk{We observe the storage challenges related to all those disaggregated systems, and pose an analysis in Section~\ref{sec:analysis}.}

\subsubsection{Request prioritization.} The intra-phase request prioritization also significantly impacts performance, particularly in terms of latency. FastServe~\cite{wu2023fastserve} uses preemptive scheduling to minimize the total latency, giving shorter requests a higher priority to minimize the total latency. \cite{sheng2024fairness} defines the fairness between users and treats the requests from different users with different priorities. \cite{fu2024efficientLLMScheduling} proposes prioritizing requests with shorter estimated computation time, aiming to reduce the average queuing time. 

\subsubsection{Throughput-oriented optimization.} Those works mainly focus on enlarging the batch size or improving the hardware utilization, and are not engaged in latency objectives or SLOs. The paged KV cache management in vLLM~\cite{kwon2023vllm} significantly improves the throughput of the LLM serving system, by removing the fragments caused by incontiguous storage. NanoFlow\cite{zhu2024nanoflow} proposes fine-grained intra-device parallelism, offering operational-level scheduling to maximize hardware efficiency. Other works~\cite{tensorrt-llm, flashattention, hong2024flashdecoding, deepspeed} investigate the faster implementation of GPU kernels, which also contribute to the throughput enhancement. 
\section{Analysis and Motivation}\label{sec:analysis}
\hk{The advantage of disaggregated computation lies in its ability to eliminate the latency interference between \textit{prefill} and \textit{decode} phases, thereby enabling a simplified and phase-specific scheduling strategy, as extensively investigated in prior disaggregated system research~\cite{patel2023splitwise, zhong2024distserve, qin2024mooncake}}. In this section, we quantify the drawbacks of disaggregated storage from the four aforementioned aspects: 1) storage imbalance, 2) KV cache transfer overhead, 3) non-negligible worker adjustment overhead, and 4) weight replica, thereby demonstrating that unified storage can be a beneficial solution.

\subsection{Storage Imbalance}
\begin{figure}
    \centering
    \includegraphics[width=0.49\linewidth]{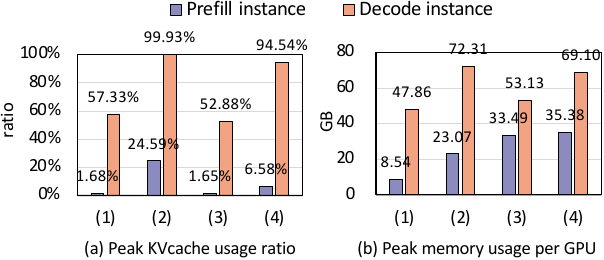}
    \caption{Storage imbalance between \textit{prefill} instance and \textit{decode} instance. (1) and (2) are measured with Llama3-8B, on ShareGPT and LongBench datasets, respectively. We set TP=2/1 for \textit{prefill}/\textit{decode} instances. (3) and (4) are measured with Llama3-70B, on ShareGPT and LongBench datasets, respectively. We set TP=4 for both instances.}
    \label{fig:imbalance}
\end{figure}

The whole KV cache is stored in the \textit{decode} instance for disaggregated systems. Meanwhile, the \textit{prefill} instance only holds the initial KV cache generated by the \textit{prefill} phase, and transfers it immediately after \textit{prefill} phase finishes. Such a mechanism leads to the KV cache storage imbalance between the two instances. We illustrate the imbalance on typical benchmarks in Figure~\ref{fig:imbalance}.
At most, only 25\% of the KV cache is occupied for the \textit{prefill} instance, while all of the KV cache can be utilized for the \textit{decode} instance. Although DistServe~\cite{zhong2024distserve} proposes to use a pull-based KV cache transfer for the \textit{decode} instance to alleviate the imbalance, only a small proportion of KV cache stays on the \textit{prefill} instance. As a result, up to 89.33\% of the GPU memory space will be wasted on the \textit{prefill} instance and the disaggregated storage disables the utilization of such wasted memory space.

\begin{figure}
    \centering
    \includegraphics[width=0.49\linewidth]{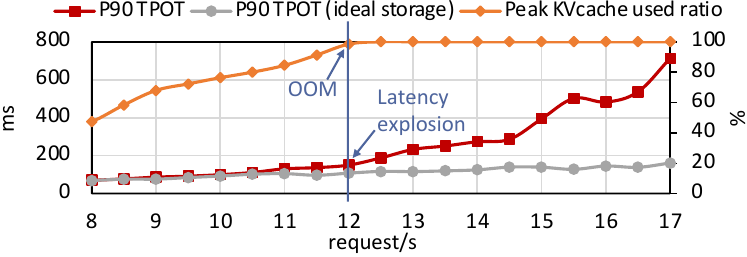}
    \caption{The consistency between storage shortage and latency explosion. The data is measured on an A100 40GB GPU, and the reported request rate is per GPU. P90 TPOT increases drastically when the storage space is exhausted. For comparison, we draw the P90 TPOT curve under the ideal storage.}
    \label{fig:explosion}
\end{figure}

Such inefficient storage causes the \textit{decode} instance to run out of memory (OOM) early, leading to latency explosion due to waiting or preemption. We present such causality in Figure~\ref{fig:explosion} by showing the consistency between storage shortage and TPOT explosion. The storage shortage leads to 3$\times$ higher TPOT, significantly harming the serving efficiency.

\subsection{KV Cache Transfer Overhead}
Since the \textit{prefill} \textit{decode} phases are disaggregated to different GPUs, the \textit{prefill} instance needs to transfer the generated KV cache to the \textit{decode} instance for computation. DistServe~\cite{zhong2024distserve} and Splitwise~\cite{patel2023splitwise} report that the KV cache transfer overhead is negligible compared to the end-to-end latency, based on the claim that the transfer happens only once, but the token generation repeats until the <\textit{eos}> token or the maximum token length is met. 
Splitwise~\cite{patel2023splitwise} introduces the layer-wise transfer to overlap the KV cache transfer overhead and the \textit{prefill} computation. 
\hk{Although KV cache transfer latency does not constitute the primary bottleneck in disaggregated systems, it nevertheless introduces non-trivial complexity to system design, including communication pattern and parallelism strategy.  
Besides, we observe that the KV cache transfer overhead is comparable to the latency of a single token generation, indicating that the second output token suffers a multiple times longer delay than the others. }

\subsection{Resource Adjustment Overhead}
\hk{Most disaggregated systems are coupled with resource adjustment mechanisms. While unified systems naturally support resource adjustment between the \textit{prefill} and \textit{decode} phases, disaggregated systems require additional design to facilitate such adjustments due to the disaggregation of resources.} There are three ways to perform resource adjustment between phases: 
1) periodic replanning in DistServe~\cite{zhong2024distserve} that generates the new parallelism pattern to adjust the resource partition between \textit{prefill} and \textit{decode} phases, 2) mixed machine pool in Splitwise~\cite{patel2023splitwise}
is maintained to perform mixed batching as needed, and 3) instance flip in TetriInfer~\cite{hu2024tetriinfer} that flips an instance running low workloads. 

We identify that the adjustment methods in disaggregated systems suffer from coarse-grained control and high overhead. All of 1), 2) and 3) make adjustments at the granularity of one GPU, together with its computational capacity and bandwidth. DistServe reports an overhead of several minutes with 1), including reloading weights, which is overwhelming in the real-time serving scenarios. With 2) the mixed batching of requests from different phases tends to cause latency interference again. The overhead of 3) comes from waiting for the existing queue to finish, which brings under-utilization as the batch size shrinks. The system chooses to drain the existing requests as it suffers more from frequent KV cache migration.

\subsection{Replicated Weights}
For disaggregated systems, both \textit{prefill} and \textit{decode} instances hold a copy of complete model weights on corresponding GPUs. The replicated weights disable the flexible deployment for disaggregated systems. For example, deploying Llama3-8B instances needs at least 2 GPUs. Two servers equipped with 16 A100 80GB GPUs can not hold a pair of \textit{prefill} and \textit{decode} instances for Llama3.1-405B. But a unified system still has 470GB of space left with such two servers after deployment. Thus, the disaggregated systems demand far more GPUs due to the replicated weights. However, the replicated instances necessitate the request rate to be large enough to fully utilize the GPU capacity, which is not always true in practice. 

\section{System Design}\label{sec:system}
Based on the analysis in Section~\ref{sec:analysis}, the combination of disaggregated computation and unified storage takes the dominant advantages of the disaggregated and the unified systems. Therefore, we build an LLM serving system, \nickname, characterized by disaggregated computation and unified storage. In this section, we describe the system design of \nickname. 

\begin{figure}[t]
    \centering
    \includegraphics[width=0.49\linewidth]{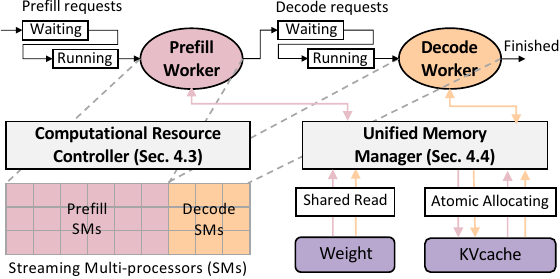}
    \caption{System overview of \nickname.}
    \label{fig:overview}
\end{figure}

\subsection{Overview}

Figure~\ref{fig:overview} presents the system overview. The disaggregated computation of each phase is denoted as a worker, and the \textit{prefill} and \textit{decode} workers handle \textit{prefill} and \textit{decode} requests, respectively. Each worker maintains its own waiting and running queue for arrived requests. On top of the two workers, the disaggregated computation is achieved via a computational resource controller, which decides the computational resource partition between the two workers, and performs the real-time switching as needed. For the storage part, \nickname manages the system's memory utilization via a unified memory manager, which controls both workers' memory access of weight and KV cache.

\subsection{\textit{Prefill} and \textit{Decode} Workers}
Same as the disaggregated designs, the \textit{prefill} and \textit{decode} workers are separated in \nickname, each owning a single process and performing asynchronous computation independently. For a single request, it is first stacked into the waiting queue of the \textit{prefill} worker upon arrival. The \textit{prefill} worker schedules the request for running and performs the \textit{prefill} iteration. Meanwhile, the generated $K, V$ projection of the request is written into the KV cache. When \textit{prefill} phase finishes, the request is stacked into the waiting queue of \textit{decode} worker. The \textit{decode} worker schedules it for running and performs the \textit{decode} iteration repeatedly until it is finished. At each \textit{decode} iteration, the KV cache of the request is updated. 

\begin{figure}
    \centering
    \includegraphics[width=0.49\linewidth]{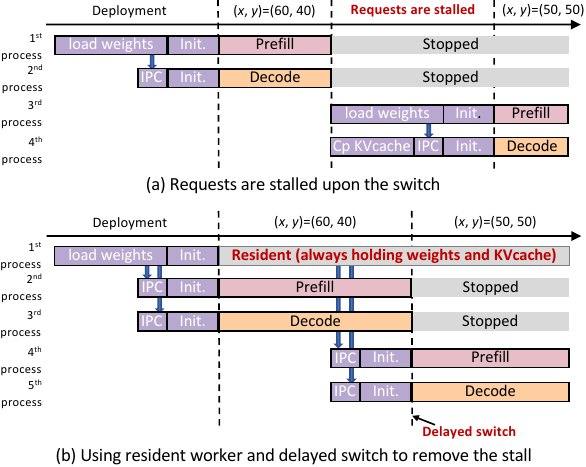}
    \caption{Low-overhead switch mechanism in \nickname.}
    \label{fig:backup}
\end{figure}

\subsection{Computational Resource Controller} 
\textbf{Disaggregated computation.} 
We implement the disaggregated computation based on the Multi-Process Service (MPS) CUDA application interface~\cite{nvidia_mps}. The MPS allows dispatching a certain number of SMs to a process, thereby achieving the computational resource partition from the SM level. \nickname receives a $(x, y)$ configuration, where $x$ and $y$ denote the percentage of the total SMs dispatched to the \textit{prefill} and \textit{decode} processes via MPS, respectively.

\noindent \textbf{Resource adjustment: challenge.} Then, we propose a lightweight resource adjustment mechanism.
As mentioned in Section~\ref{sec:system}, each worker is tied to a process, and a pair of processes can be assigned with only one $(x, y)$.
MPS does not support adjusting $(x, y)$ of the existing processes.
Thus, adjusting $(x, y)$ needs to reinvoke the MPS interface, thereby bringing overhead from the processes' switch.
We need to wait for the \textit{prefill} worker and \textit{decode} worker to finish their running iteration and store the generated token.
Then, we kill the two processes.
Finally, we renew two processes using the MPS interface as the new \textit{prefill} worker and \textit{decode} worker with new $(x, y)$. Based on the KV cache and the next token, the new workers go on.
Thus, the overhead comes from two aspects.
First, as shown in Fig.~\ref{fig:backup}, renewing two MPS processes includes 1) loading the weights to the new \textit{prefill} worker and sharing with the new \textit{decode} worker via IPC (or vice versa), 2) copying the KV cache to the new \textit{decode} worker, and 3) initializing the engine for both new workers. Such switching overhead causes the arrived requests to be stalled until the new workers are ready. 
Second, the switching timing is also important.
Because \textit{prefill} workers and \textit{decode} workers have different processes, we need to insert a synchronous operation to stall the early finished worker, but the stalling behavior leads to an idle period for request processing.  

\noindent \textbf{Resource adjustment: solution.} To solve the first overhead, we introduce a resident process to consistently hold the weights and KV cache during serving, avoiding the repeated loading of weights and copying of KV cache, as shown in Fig.~\ref{fig:backup}(b).
Then we share the pointer of the allocated space to the \textit{prefill} process and \textit{decode} process via the inter-process communication (IPC)~\cite{nvidia_ipc} so that the \textit{prefill} and \textit{decode} workers can have access to the weights and KV cache through the pointer.
In that way, the memory space of weights and KV cache is not released with the stopped worker, and the new workers can access the storage through IPC at negligible time cost. To hide the latency of IPC and initialization, \nickname conducts the delayed switching, running under the new $(x, y)$ only when the preparation step finishes. 

To solve the second overhead, we propose asynchronous switching.
We can renew two MPS processes directly and only kill the worker who has finished its iteration.
Such an asynchronous behavior ensures there are always \textit{prefill} and \textit{decode} processes running in the system. Situations can arise temporally with an old \textit{prefill} worker and a new \textit{decode} worker running together, and vice versa, which are allowed in the system. 
This is attributed to the fact that MPS allows the resource percentage of all running processes to be larger than 100 percent, where each process will compete for the resources for a short while.

\subsection{Unified Memory Manager}
\begin{figure}
    \centering
    \includegraphics[width=0.49\linewidth]{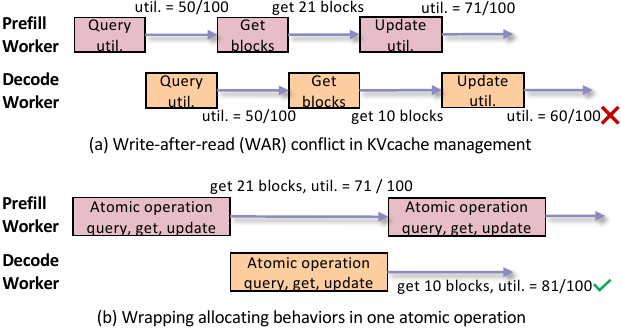}
    \caption{The atomic allocating operation avoids the WAR conflict.}
    \label{fig:atomic}
\end{figure}

Two types of memory access are involved in \nickname, \textit{i.e.}, reading model weights and accessing KV cache. 
The unified memory manager can straightforwardly support the model weights since they are read-only.
Thus, we mainly focus on the KV cache accesses.
We follow vLLM~\cite{kwon2023vllm} to use a paged storage for the KV cache, and the KV cache is accessed through the block table index. Once the block table index is determined, the access of the KV cache can be conducted without conflicts. However, conflicts happen when the \textit{prefill} worker and the \textit{decode} worker asynchronously allocate the KV cache. The \textit{prefill} worker allocates the KV cache when a layer-wise computation is finished, and the \textit{decode} worker allocates the KV cache for every layer in the PagedAttention~\cite{kwon2023vllm} computation. The asynchronous manner of the two workers potentially brings write-after-read (WAR) conflicts to memory management. As illustrated in Figure~\ref{fig:atomic}, the allocating contains three steps: querying the memory utilization to see if there are free blocks, getting the blocks for KV cache storage, and updating the memory utilization. The WAR conflict happens when one worker immediately updates the utilization after it is queried. The result is that the querying worker updates the wrong memory utilization ratio based on the queried value. To address the challenge, we introduce the atomic operation for KV cache block allocating, \textit{i.e.}, the memory utilization is locked until the update step finishes. Such an atomic allocating pattern ensures that memory utilization is correct against the two asynchronous processes.

\subsection{Multi-GPU Scheme}
\begin{figure}
    \centering
    \includegraphics[width=0.49\linewidth]{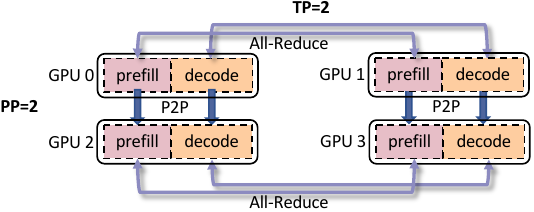}
    \caption{A TP=2 \& PP=2 example with multi-GPU scheme.}
    \label{fig:multi-gpu}
\end{figure}

\textbf{Scheme.} \nickname supports serving LLMs across multiple GPUs, and the scheme is illustrated in Figure~\ref{fig:multi-gpu}. Both \textit{prefill} and \textit{decode} workers reside on a single GPU, carrying their computation asynchronously. Thus, the inter-GPU communication happens asynchronously for the two workers. Specifically, within a TP group (\textit{e.g.}, GPU 0 and GPU 1 in Figure~\ref{fig:multi-gpu}), all the \textit{prefill} workers communicate with each other to conduct the all-reduce operation. And within a PP group (\textit{e.g.}, GPU 0 and GPU 2 in Figure~\ref{fig:multi-gpu}), the \textit{prefill} worker sends the activation to the \textit{prefill} worker on the next GPU. The \textit{decode} workers behave the same. Given a parallelism setting, the amount of data transferred remains the same as in a unified system. The limitation of such a scheme is the parallel patterns have to be the same for \textit{prefill} and \textit{decode} phases. However, we discuss in the following that the limitation is not an influential factor for serving performance. 

\noindent \textbf{Discussion. }
In a disaggregated system, the parallelism decoupling denotes that the \textit{prefill} and \textit{decode} workers can employ different parallel strategies (\textit{e.g.}, TP=4 and PP=1 for the \textit{prefill} worker while TP=2 and PP=2 for the \textit{decode} worker).
We identify that the ratio of the number of GPUs of \textit{prefill} workers and \textit{decode} workers plays a more important role than different parallelism patterns.



\hk{Taking the commonly used TP and PP in inference as examples, their impacts on requests during the \textit{prefill} and \textit{decode} phases are similar. TP partitions a single request across multiple devices for parallel computation, which helps reduce request-level latency but introduces higher communication overhead. In contrast, PP enables parallel computation among different layers of distinct requests, which has lower communication overhead, thereby enabling the throughput to scale up with GPUs. Leveraging the insights from DistServe~\cite{zhong2024distserve}, TP is more beneficial for latency reduction under low request rates, whereas PP is required to enhance throughput under high request rates, which is a conclusion applicable to both \textit{prefill} and \textit{decode} phases.}

\begin{figure}
    \centering
    \includegraphics[width=0.49\linewidth]{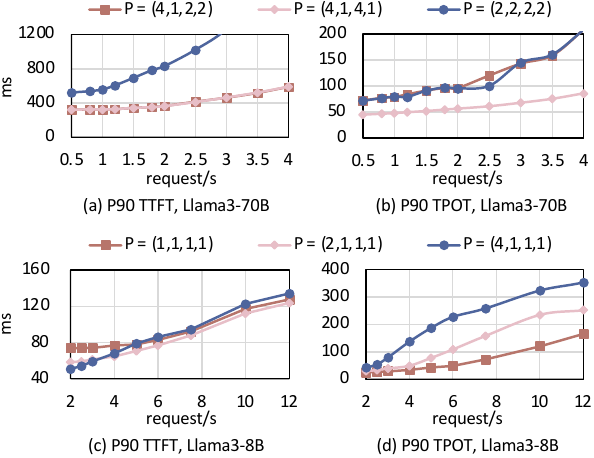}
    \caption{Performance of DistServe under different parallelism settings. The reported request rate is per GPU.}
    \vspace{-0.0em}
    \label{fig:parallelism}
\end{figure}

As a supplement, we conduct experiments to validate the conclusion. 
We use $P=(t^\text{p}, t^\text{d}, p^\text{p}, p^\text{d})$ to represent the TP/PP degrees for \textit{prefill} ($t^\text{p}/p^\text{p}$) and \textit{decode}  ($t^\text{d}/p^\text{d}$) phases, respectively.
As shown in Figure~\ref{fig:parallelism}, 
we test the performance of DistServe with Llama3-8B and Llama3-70B on the ShareGPT dataset.
In Figure~\ref{fig:parallelism}(a) and (b), for the 70B model, we observe that using $P=(4,1,4,1)$ continuously outperforms other settings, indicating tensor parallelism is better than pipeline parallelism for both phases.
It demonstrates that when we fix the GPU number of \textit{prefill} and \textit{decode} instances, decoupled parallelism cannot obtain performance benefits.
Furthermore, Figure~\ref{fig:parallelism}(c) and (d) depict that different ratios of GPU number of \textit{prefill} workers and \textit{decode} instances have a significant impact on performance.
\section{Dynamic Adjusting Method}

For latency-sensitive serving tasks, the goal of the serving system is to serve more requests adhering to the given service-level objectives (SLOs). For LLM serving, the SLOs refer to the latency constraints on TTFT and TPOT. Therefore, the resource partitioning between the \textit{prefill} and \textit{decode} workers (\textit{i.e.}, $(x, y)$) should consider SLOs. Besides, the \textit{prefill} and \textit{decode} workloads change over time. A fixed resource partitioning $(x, y)$ often leads to SLO violation, which arises from the mismatch between the dispatched resources and the dynamic workload demands. To address the issue, \nickname enables an SLO-aware method to dynamically adjust the resource partitioning $(x, y)$ following the workload changes. We first introduce a resident worker and a delayed switch mechanism to reduce the adjustment overhead. Based on that, we further design an SLO-aware algorithm to adjust the resource partitioning, which takes the given SLOs as inputs. 

\noindent \textbf{Algorithm.}
\hk{The proposed SLO-aware algorithm employs a feedback mechanism to adjust computational resources dynamically, implicitly accounting for the impacts of both input variation and model selection.} The details are illustrated in Algorithm~\ref{alg:scheduling} and the inputs are outlined at the beginning. We conduct the iteration-level adjustment periodically based on a $window\_size$, and we collect the observed TTFT and TPOT within the window to update models for TTFT/TPOT estimation (lines 1-6). The algorithm directly returns the current partition when both TTFT and TPOT fail the SLOs, as there is no space for adjustment (line 19). Otherwise, we increase the SM ratio step by step for the phase with the failed latency and assess if the updated ratio satisfies the SLO (lines 9-10, 14-15). \hk{Given that $x, y \leq 100$, the increase in one ratio can be replaced by the reduction in the other if the increase leads to overflow.} After that, we use the normalized SM ratio $x'$ or $y'$ to estimate TTFT or TPOT, considering that the two phases share 100\% of the computational resource in total (lines 11, 16). The configuration \textit{window\_size} controls the adjustment frequency, and \textit{max\_step} and \textit{step\_size} decide the learning rate. 

\begin{algorithm}[t]
    \caption{SLO-aware adjusting algorithm} 
    \label{alg:scheduling}
    \begin{flushleft}
    \textbf{Input:} iter \#iteration index, $(x_0, y_0)$ \#current partition, $(S^{\text{p}}, S^{\text{d}})$ \#user-defined SLOs, p$^{\text{SLO}}$ \#satisfied percentage defined in SLOs, \textit{window\_size}, \textit{max\_step}, \textit{step\_size} \#algorithm configurations
    \end{flushleft}
    \begin{algorithmic}[1]
        \If{iter \% \textit{window\_size} > 0}
            \State return $(x_0, y_0)$
        \EndIf
        \State $(x, y) \gets (x_0, y_0)$ 
        \State TTFT, TPOT $\gets$ get\_observed\_latency(p$^{\text{SLO}}$)
        \State update\_estimate\_model($x$, $y$, TTFT, TPOT)
        \State step $\gets 0$
        \If {TTFT SLO fails}
            \While {step < \textit{max\_step} and estimate\_ttft($x'$) $> S^{\text{p}}$}
                \State $(x, y) \gets$ increase\_x\_ratio($x, y$, \textit{step\_size})
                \State $x' \gets x / (x + y)$
            \EndWhile
        \ElsIf {TPOT SLO fails}
            \While {step < \textit{max\_step} and estimate\_tpot($y'$) $> S^{\text{d}}$}
                \State $(x, y) \gets$ increase\_y\_ratio($x, y$, \textit{step\_size})
                \State $y' \gets y / (x + y)$
            \EndWhile
        \Else
            \State return $(x_0, y_0)$
        \EndIf
        \State return $(x, y)$
    \end{algorithmic}
    \vspace{-0.em}
\end{algorithm}

\noindent \textbf{Latency modeling.} 
At lines 9 and 15 in Algorithm~\ref{alg:scheduling}, we need to model the relationship between TTFT and TPOT against the accessible SM ratio $x$ and $y$, respectively. Since the two workers asynchronously handle requests, we model TTFT and TPOT separately. The TTFT of a single request consists of two parts of latency, \textit{i.e.}, waiting latency, and \textit{prefill} latency. On the other hand, we assume the requests are immediately batched for running upon arrival at the \textit{decode} worker, hence excluding the waiting latency from TPOT. The more SMs a process can access, the more computational resources it can utilize in parallel, leading to a proportionally growing processing rate. Thus, we simply model the processing latency as 
\begin{align}\label{eq:linear}
l_{x} = \frac{100}{x} \times l_{100}.
\end{align}
Here, $l_{x}$ represents the processing latency under the SM ratio $x$ for both \textit{prefill} and \textit{decode} phases. For TTFT, we apply an M/M/1~\cite{shortle2018queueingtheory} queuing model to formulate the sum of \textit{waiting} and \textit{prefill} latencies as follows
\begin{align}\label{eq:wait}
w = \frac{1}{\mu_x - r},
\end{align}
where $r$ and $\mu_x$ denote the given input request rate and the \textit{prefill} processing rate under SM ratio $x$, respectively. $w$ is the sum of the waiting latency and the processing latency. Based on Eq.~\ref{eq:linear}, we have $\mu_x = 1 / l_{x} \propto x$, Eq.~\ref{eq:wait} is further written as 
\begin{align}
w \propto \frac{1}{x - \lambda},~\lambda = 100 r l_{100}.
\end{align}
Therefore, the relationships between TTFT and $x$, TPOT and $y$ are expressed as
\begin{align}\label{eq:final}
TTFT_x = a_1 \frac{1}{x - \lambda} + b_1, ~TPOT_y = a_2 \frac{1}{y} + b_2.
\end{align}

\begin{figure}
    \centering
    \includegraphics[width=0.6\linewidth]{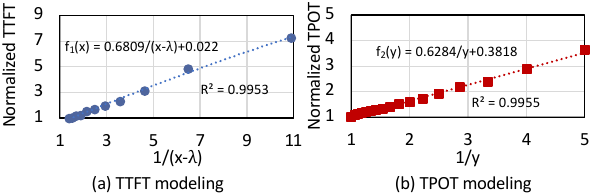}
    \caption{The inverse linear relationship of TTFT/TPOT against the SM accessible ratio. $R^2$ denotes the coefficient of determination, being closer to 1 is better. Data is collected with Llama3-8B and ShareGPT~\cite{sharegpt2023sharegpt}.}
    \label{fig:modeling}
\end{figure}

The parameters $a_1, a_2, b_1, \lambda, b_2$ are learnable and computed based on the real-time observation of TTFT and TPOT. Note that we introduce the constant terms $b_1, b_2$ to model the constant overhead in the latency. Given $(x, y)$, we utilize Eq.~\ref{eq:final} to estimate TTFT and TPOT, so that we can quantify the impact of adjusting $(x, y)$ during serving. Figure~\ref{fig:modeling} validates the proposed latency modeling, where (a) shows $TTFT_x$ satisfying the linear relationship with $1/(x - \lambda)$, and (b) shows $TPOT_y$ satisfying the linear relationship with $1/y$. 

\section{Implementation}\label{sec:impl}




\hk{We build \nickname on both DistServe~\cite{zhong2024distserve} and SGLang~\cite{zheng2023sglang} codebases to support instance-scale deployment, incorporating the system modifications described in Section~\ref{sec:system} to enable disaggregated computation and unified storage. For the cluster environment, we modify the NVIDIA Dynamo~\cite{dynamo} frontend and use vLLM~\cite{kwon2023vllm} as the backend. } 

\noindent \textbf{Implementation based on DistServe.}
We modify the grouped-query attention (GQA) with the optimized kernels from FlashAttention~\cite{flashattention2} for both phases. Besides, the first two general matrix multiplications (GEMMs) in the gated FFN are fused into one single GEMM. To support the Llama3.1 series~\cite{dubey2024llama} model, we also modify the RoPE kernel. Based on those modifications, \nickname achieves comparable performance with state-of-the-art inference engines at the operator level. 
Note that we use all those optimized GPU kernels for DistServe as the baseline in the evaluation section. 

To support uneven pipeline parallelism when the number of layers is not divisible by the parallelism degree (\textit{e.g.,} Llama3.1-405B with 126 layers), we use the even pipeline parallelism so that the model can be deployed across 4 nodes with PP=4. Specifically, we dispatch the last two layers to the nodes in a round-robin manner. 

\nickname simply utilizes the first-come-first-serve (FCFS) scheduling for request processing, and the chunked-prefill mechanism is not supported for the \textit{prefill} phase with the DistServe implementation. For the \textit{decode} phase, we set a maximum batch size to control the batched request number. 

\noindent \textbf{Implementation based on SGLang.}
For the SGLang implementation, we develop a disaggregated frontend where \textit{prefill} and \textit{decode} instances operate as separate single-program-multiple-data (SPMD) processes that maintain consistent model parallelism strategies. We implement a state-copy mechanism between the \textit{prefill} and \textit{decode} message queues, enabling simultaneous request reception for the two phases, thereby eliminating inter-process communication overhead.

The system also utilizes first-come-first-serve (FCFS) scheduling for request processing. For the \textit{prefill} phase, \nickname employs the chunked-prefill mechanism to control the batched token number, avoiding overwhelming memory consumption brought by activation or prolonged batch processing latency. For the \textit{decode} phase, we implement batch size controlling and incorporate the CPU-GPU overlap scheduling following SGLang.

\begin{figure}
    \centering
    \includegraphics[width=0.4\linewidth]{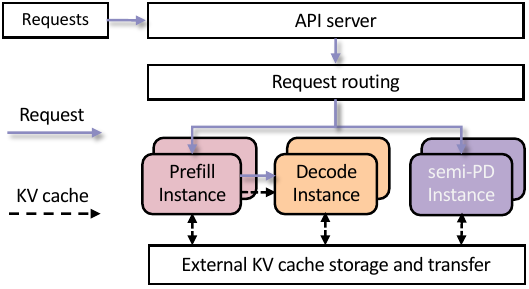}
    \caption{\nickname instance applied in cluster-scale serving.}
    \label{fig:cluster}
\end{figure}

\hk{\noindent \textbf{Scalable Deployment.}
In single-instance deployment, \nickname can be used for complete deployment of a single model, particularly suited for private or local deployment environments. The disaggregated computation in \nickname effectively eliminates latency interference while the unified storage substantially improves GPU utilization efficiency, thereby enabling reductions in deployment costs.}

\hk{In cluster environments, \nickname is treated as an instance selection together with \textit{prefill} and \textit{decode} instances, as illustrated in Figure~\ref{fig:cluster}. The requests are routed to the \textit{prefill} instances or the \nickname instances for the \textit{prefill} phase processing. For the former case, requests are transferred from the \textit{prefill} instance to the \textit{decode} instance for computation, while for the latter, requests complete the \textit{decode} phase computation within the same \nickname instance. The request router dynamically routes requests between the disaggregated and \nickname instances. Specifically, \nickname is integrated into NVIDIA Dynamo, and on top of that, a router that collaboratively considers KV cache hit rate, GPU utilization, and waiting queue depth is designed to determine where the arrived requests go. }
\section{Evaluation}


\hk{The evaluation of \nickname comprises two environments: 1) instance-scale LLM serving and 2) cluster-scale LLM serving. For both environments,} we evaluate \nickname on various models and datasets, comparing the P90/P99 TTFT and TPOT to the baselines. We observe that as the input request rate grows, \nickname maintains a relatively low latency for both TTFT and TPOT. \hk{Notably, \nickname reduces the average end-to-end latency per request by 1.27-2.58$\times$ when serving DeepSeek series models. Also, considering the SLO attainment, \nickname serves \hk{1.55-1.72}$\times$ more requests than the SOTA systems adhering to the SLOs on Llama series models.} We also conduct experiments to evaluate the proposed dynamic adjustment algorithm and the low-overhead switching mechanism. \hk{Note that all request rates are per GPU in this section.}

\subsection{Setup}
\subsubsection{Testbed} 
\hk{The experiments involve four platforms. \textit{Server A}: one node with eight NVIDIA A100 80GB SXM4 GPUs connected with pairwise NVLink~\cite{nvlink2024}. \textit{Server B}: four nodes with each equipped by eight NVIDIA A100 40GB SXM4 GPUs. The cross-node bandwidth is 200 Gbps. \textit{Server C}: one node with eight NVIDIA H200 141GB SXM5 GPUs connected with pairwise NVLink~\cite{nvlink2024}. \textit{Server D}: one node with eight NVIDIA A800 80GB SXM4 GPUs connected with pairwise NVLink~\cite{nvlink2024}.} 
The software environment includes CUDA 12.1~\cite{nvidia2024cuda}, NCCL 2.18~\cite{awan2016efficient}, PyTorch 2.3.0~\cite{paszke2019pytorch}, etc. 

\subsubsection{Model} As listed in Table~\ref{tab:workloads}, for the implementation based on DistServe, we evaluate \nickname using the Llama3/3.1~\cite{meta2024llama3, dubey2024llama} series models, with the model sizes of 8B, 70B and 405B. We deploy the 8B and 70B models on \textit{Server A} while the 405B model on \textit{Server B}. All parameters used for experiments are in FP16 precision. As demonstrated in Figure~\ref{fig:parallelism}, considering both phases, TP=1 and TP=4 are respectively the best choices for the 8B and 70B models under latency constraints. Thus, we adopt TP=1 and TP=4 in the 8B and 70B evaluations. For the 405B model, we apply TP=8 and PP=4, with PP across different nodes. 

\hk{For the implementation based on SGLang, we evaluate \nickname using the DeepSeek~\cite{deepseek-v2, deepseek-v3} series models. The DeepSeek-V2-Lite model has a parameter size of 16B, and is directly deployed on a single A100/A800 80GB GPU. The DeepSeek-V3 model has a parameter size of 671B and is deployed on \textit{Server C} with TP=8 and FP8 precision. }

\subsubsection{Dataset} 
\hk{We choose representative workloads including chatbot, long context and math. The benchmarked datasets are ShareGPT~\cite{sharegpt2023sharegpt}, LongBench~\cite{bai2023longbench}, and MATH-500~\cite{math500}, respectively. For each workload, we sample from suitable datasets and generate request arrival times using the Poisson distribution to emulate the real-world workloads. The datasets used for each model are illustrated in Table~\ref{tab:workloads}.}

\hk{Regarding cluster-scale serving, we evaluate \nickname on heterogeneous workloads to emulate the real-world scenario. For that purpose, we create a synthetic dataset consisting of 95\% requests sampled from ShareGPT to represent the regular workloads, and 5\% irregular requests owning long input length ($\sim$4k compared to ShareGPT's 251 on average). }

\subsubsection{Baseline}
\hk{For the Llama series models, we compare \nickname against DistServe and vLLM, and for the instance-scale serving of DeepSeek series models, we compare \nickname against SGLang. Furthermore, NVIDIA Dynamo is taken as the baseline for cluster-scale evaluation. The baselines are detailed as follows.} 
\begin{itemize}
    \item \textbf{DistServe}~\cite{zhong2024distserve}: One of the few disaggregated systems that release open-source codes. \nickname uses the same GPU kernels and the same scheduling settings as DistServe. Specifically, \textit{max\_batch\_size} is set to 512. \hk{When evaluated in instance-scale serving, DistServe is set to use one \textit{prefill} instance and one \textit{decode} instance (\textit{i.e.}, 1P1D). Therefore, the occupied GPU number is doubled for DistServe in Table~\ref{tab:workloads}.}
    \item \textbf{vLLM (v0.6.0)}~\cite{kwon2023vllm}: A unified system accompanied by multiple open-source optimizations, we compare with its default implementation (\textbf{vLLM-D}) and SplitFuse implementation (\textbf{vLLM-S}). We set \textit{max\_batch\_size}=512 for both vLLM-D and vLLM-S, and use \textit{chunk\_size}=1024 for vLLM-S.
    \hk{\item \textbf{SGLang (v0.4.4)}~\cite{zheng2023sglang}: Another unified system with high-throughput optimization, providing support for DeepSeek-V3 serving upon the model's release. } 
    \hk{\item \textbf{Dynamo (v0.1.1)}~\cite{dynamo}: A cluster-scale distributed inference serving system, featuring \textit{prefill} and \textit{decode} disaggregation, currently only supporting vLLM~\cite{kwon2023vllm} as backend. When serving DeepSeek-V2-Lite in evaluation, Dynamo uses two \textit{prefill} instances and six \textit{decode} instances (\textit{i.e.}, 2P6D), while we replace one of the \textit{prefill} instances and three of the \textit{decode} instances with four \nickname instances (denoted as 1P3D4S). }
\end{itemize}

\begin{table}
  \caption{Platform, benchmark, and model settings.}
  \centering
  \label{tab:workloads}
  \resizebox{0.98\textwidth}{!}
  {
  \begin{tabular}{cccccc}
    \toprule
    Enviroment & Platform & Codebase & Model & Parallelism & Dataset \\
    \midrule
    \multirow{5}[0]{*}{Instance} & 1$\times$A100 80GB SXM4 GPU (\textit{Server A}) & \multirow{3}[0]{*}{DistServe} & Llama3-8B & - & \multirow{3}[0]{*}{ShareGPT, LongBench} \\
        & 4$\times$A100 80GB SXM4 GPUs (\textit{Server A})
    &  & Llama3-70B & TP=4 &  \\
        & 32$\times$A100 40GB SXM4 GPUs (\textit{Server B})
    &  & Llama3.1-405B & TP=8, PP=4 &  \\
    \cmidrule(lr){3-6}
        & 1$\times$A100 80GB SXM4 GPU (\textit{Server A})
    & \multirow{2}[0]{*}{SGLang} & DeepSeek-V2-Lite & - & ShareGPT \\
        & 8$\times$H200 141GB SXM5 GPUs (\textit{Server C})
    &  & DeepSeek-V3 & TP=8 & MATH-500 \\ 
    \midrule
    Cluster & 8$\times$A800 80GB SXM4 GPUs (\textit{Server D})
    & Dynamo & DeepSeek-V2-Lite & - & Synthetic \\
    
    \bottomrule
  \end{tabular}
  }
\end{table}
\vspace{.0em}

\begin{figure*}[t]
    \centering
    \includegraphics[width=0.98\linewidth]{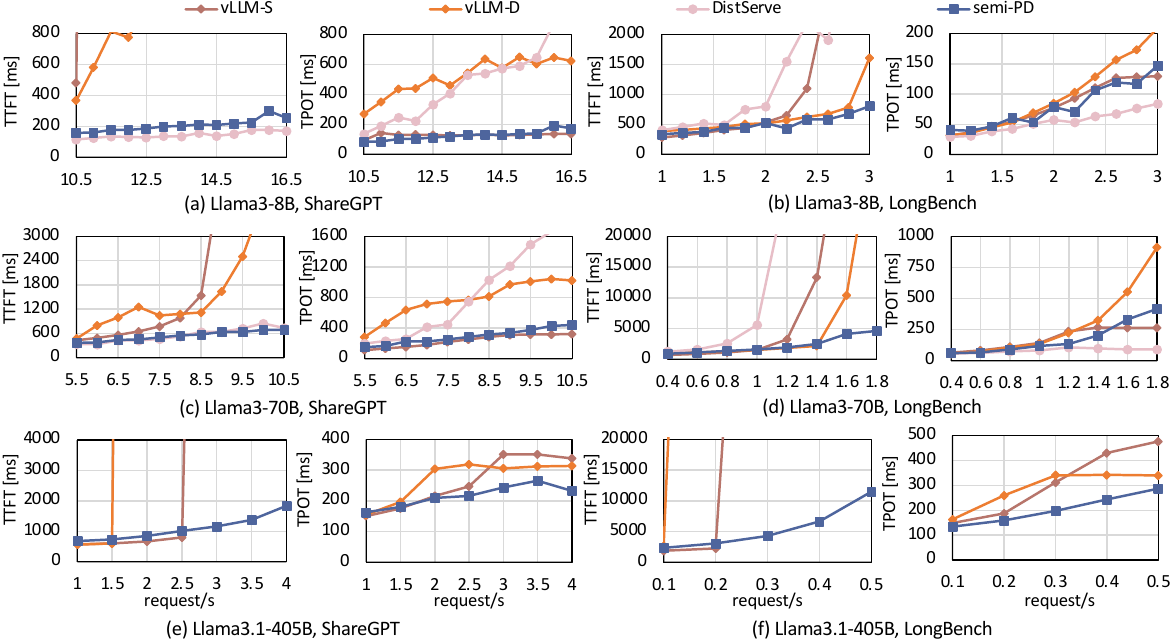}
    \vspace{-0.em}
    \caption{The P90 TTFT and TPOT comparison on Llama series models (lower is better). For Llama3.1-405B, we only compare \nickname with vLLM-S and vLLM-D, as DistServe can not be deployed due to the storage limitation.}
    \vspace{-0.0em}
    \label{fig:e2e}
\end{figure*}

\begin{figure}[t]
    \centering
    \includegraphics[width=0.98\linewidth]{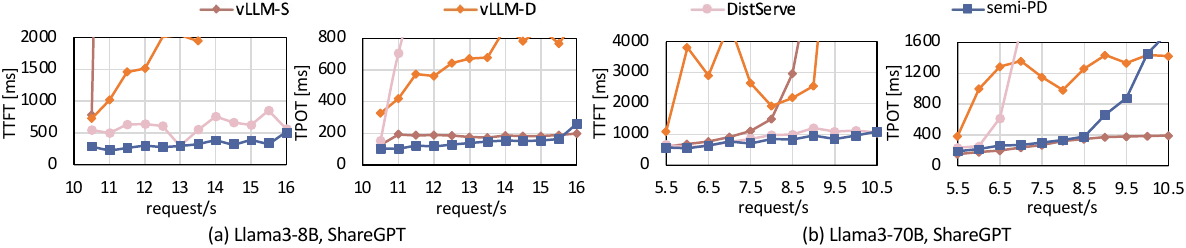}
    \vspace{-0.em}
    \caption{The P99 TTFT and TPOT comparison on Llama series models (lower is better). }
    \vspace{0.em}
    \label{fig:p99}
\end{figure}

\begin{figure}[t]
    \centering
    \includegraphics[width=0.98\linewidth]{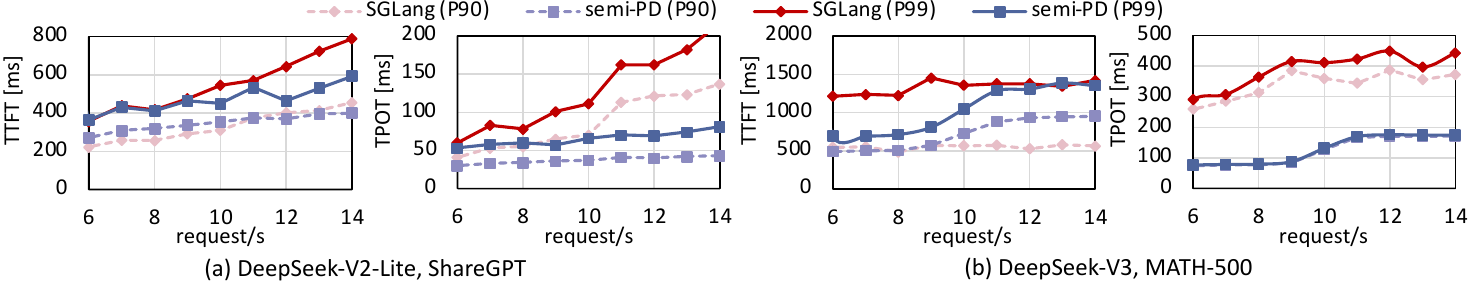}
    \vspace{-0.em}
    \caption{The P90/P99 TTFT and TPOT comparison on DeepSeek series models (lower is better). }
    \vspace{0.em}
    \label{fig:deepseek}
\end{figure}


\begin{table}[t]
  \caption{SLO settings in evaluation (TTFT/TPOT).}
  \centering
  \label{tab:slo}
  \resizebox{0.48\textwidth}{!}
  {
  \begin{tabular}{lcccccc}
    \toprule
    \multirow{2}[0]{*}{Model} & \multicolumn{2}{c}{ShareGPT} & \multicolumn{2}{c}{LongBench} \\
          &  Tight & Loose & Tight & Loose \\
    \midrule
        Llama3-8B & 0.3s/0.15s & 0.4s/0.2s & 2.25s/0.13s & 3.0s/0.18s\\
        Llama3-70B & 0.85s/0.3s & 1.1s/0.4s & 6.0s/0.3s & 8.0s/0.4s \\
    \bottomrule
  \end{tabular}
  }
\end{table}
\vspace{.0em}

\begin{figure*}[t]
    \centering
    \includegraphics[width=0.98\linewidth]{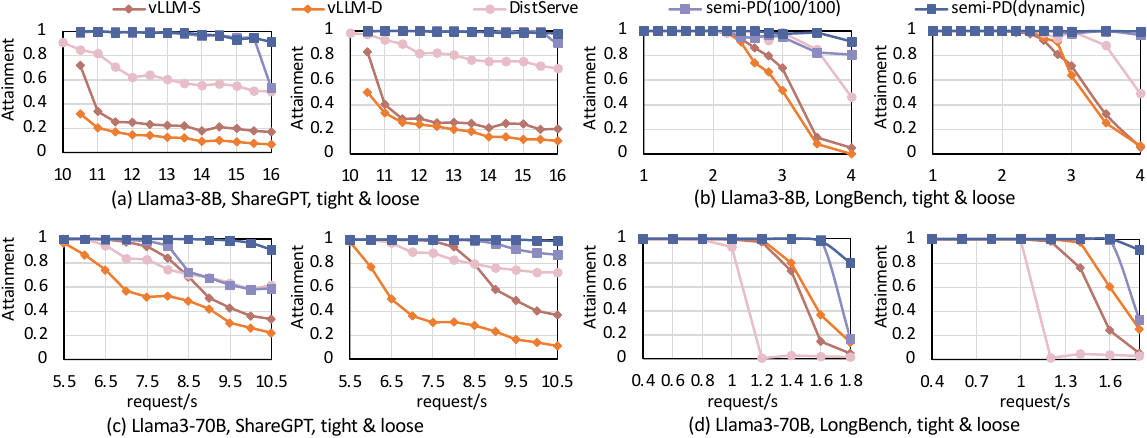}
    \vspace{-0.em}
    \caption{The SLO attainment comparison (higher is better). For (a)-(d), the tight/loose SLO results are on the left/right.}
    \vspace{-0.5em}
    \label{fig:slo}
\end{figure*}

\subsubsection{SLOs}\label{sec:slo} The SLOs are the constraints on TTFT and TPOT. The SLO settings in our experiments are in Table~\ref{tab:workloads}. We use the approximates of 7.5$\times$ and 10$\times$ of the single request latency as the tight and loose SLOs, respectively. 

\subsection{Instance-Scale Performance}
To evaluate the overall performance of instance-scale serving, we mainly compare the P90 TTFT and TPOT across all the systems. We also present the results of P99 TTFT and TPOT to further demonstrate the efficiency of the design. 

\subsubsection{P90/P99 TTFT and TPOT Comparison}

\textbf{Llama series.} As shown in Figure~\ref{fig:e2e}, other systems exhibit a sharp increase in TTFT or TPOT under high request rates, while \nickname maintains stability across both latency metrics. For example, in Figure~\ref{fig:e2e}(a), vLLM-S and vLLM-D pose a surge in the P90 TTFT, whereas DistServe increases drastically in the P90 TPOT. Meanwhile, \nickname remains relatively low TTFT and TPOT when the input request rate exceeds 16 request/s. The surge in TTFT for vLLM-S is because of the prioritization of \textit{decode} requests over \textit{prefill} requests, and the latency interference becomes significant under higher request rates. Such latency interference is reflected in all the results. The surge in TPOT for DistServe is due to storage inefficiency mentioned in Section~\ref{sec:analysis}.

Similar trends are observed on the Longbench dataset. In Figure~\ref{fig:e2e}(b) and (d), the TTFT surge mainly arises from two aspects: 1) the \textit{decode} worker of DistServe runs out of memory due to the storage imbalance, and the requests accumulate on the \textit{prefill} worker. 2) At the beginning of serving, the \textit{prefill} worker demands much more computational resources, but the disaggregated design fails to make adjustments at negligible cost. \nickname addresses the two issues and avoids the TTFT explosion. 

We compare the P99 latency results serving Llama-8B and Llama3-70B on ShareGPT. Figure~\ref{fig:p99} shows the P99 TTFT and TPOT, demonstrating a performance trend similar to Figure~\ref{fig:e2e}. The P99 metrics tend to exhibit more fluctuations across all systems. In fact, \nickname delivers better performance in P99 latency, avoiding latency explosion due to latency interference and storage imbalance. \nickname is the only one that maintains smooth TTFT and TPOT, which can handle over 11 request/s for Llama3-8B and 8 request/s for Llama3-70B. Without loss of generality, we choose to use the P90 latency as the major metric.

\hk{\noindent \textbf{DeepSeek series.} The results are depicted in Figure~\ref{fig:deepseek}. When serving DeepSeek-V2-Lite on ShareGPT, \nickname delivers lower P90/P99 TTFT and TPOT under high request rates, demonstrating the ability to serve more requests adhering to latency constraints. Although \nickname's P90 TTFT is slightly higher when serving DeepSeek-V3 on MATH-500, its P99 TTFT remains at a low level under high request rates. Notably, the TPOT of \nickname is far better than SGLang's. Furthermore, we observe that \nickname brings 1.27-2.40$\times$ and 1.49-2.58$\times$ speedups in the average end-to-end latency per request when serving DeepSeek-V2-Lite and DeepSeek-V3, respectively.}


\subsubsection{SLO Attainment}
Given SLOs, the proposed dynamic adjusting method enables the real-time resource adjustment between \textit{prefill} and \textit{decode} phases, thereby enhancing the SLO attainment. We show the attainment improvement brought by the dynamic adjusting method for evaluation. We use two different SLOs, tight and loose, as demonstrated in Section~\ref{sec:slo}, to capture a more comprehensive attainment comparison result. We take \nickname with $(x, y) = (100, 100)$ as the baseline dynamic method, which hands in the responsibility of resource partition to the CUDA scheduler, and performs uncontrollable dynamic adjustments between two workers. We denote the implementation with the proposed dynamic adjusting method as \nickname (dynamic).

Figure~\ref{fig:slo} presents the SLO attainment of all the systems. Both \nickname (100, 100) and \nickname (dynamic) outperform other baselines, maintaining the attainment at a high level as the request rate increases. \nickname (dynamic) further improves the SLO attainment compared to \nickname (100, 100), thanks to its SLO-aware algorithm as the adjustment guidance. On average, \nickname (dynamic) achieves 1.55$\times$, 1.72$\times$, 1.62$\times$, 1.11$\times$ higher request rate over vLLM-S, vLLM-D, DistServe, \nickname (100, 100), respectively, while maintaining the SLO attainment larger than 90\%.


\subsection{Cluster-Scale Performance}
\begin{figure}[t]
    \centering
    \includegraphics[width=0.55\linewidth]{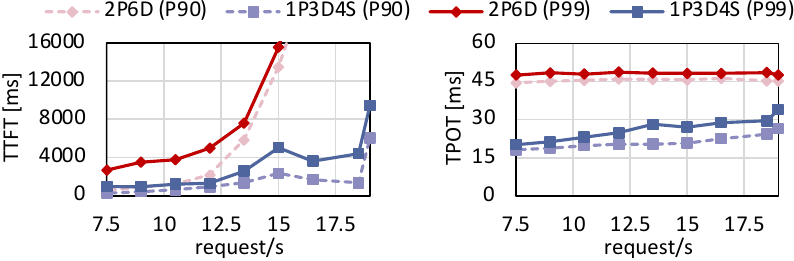}
    \vspace{-0.em}
    \caption{The cluster-scale serving latency comparison (lower is better). 2P6D denotes using pure Dynamo with two \textit{prefill} instances and six \textit{decode} instances, while 1P3D4S denotes using one \textit{prefil} instances and three \textit{decode} instances with Dynamo and four extra \nickname instances.}
    \vspace{0.em}
    \label{fig:cluster-latency}
\end{figure}

\hk{The results in Figure~\ref{fig:cluster-latency} demonstrate that the integration of \nickname instances reduces both TTFT and TPOT against heterogeneous workloads. Notably, with \nickname, the system avoids the TTFT surge under low request rates, and achieves more than 2$\times$ of the TPOT reduction. For the baseline Dynamo implementation, given eight GPUs, 2P6D is the optimal solution under the regular requests sampled from ShareGPT. However, such a static resource partition fails to keep up with sudden workload changes brought by those irregular requests. To adapt to the changes, a disaggregated system requires relatively high overhead for adjustment, as mentioned in Section~\ref{sec:analysis}. On the other hand, the \nickname instances enable low-overhead resource adaptation to workload changes, while maintaining interference-free computation between the two phases. When integrated into disaggregated systems, \nickname instance effectively functions as an elastic buffer for resource adjustment. Therefore, by replacing a portion of the instances with \nickname instances, the system can handle the heterogeneous workloads better.}

\section{Conclusion}
In this paper, we introduce \nickname, an efficient LLM serving system characterized by disaggregated computation and unified storage. \nickname follows the disaggregated system to avoid latency interference between \textit{prefill} and \textit{decode} phases, and further addresses its drawbacks of inefficient storage. Moreover, by developing a low-overhead worker switch mechanism, \nickname can adjust the resource partition between the two phases at negligible cost. Based on that, we propose an SLO-aware algorithm for \nickname, optimizing the SLO attainment by adjusting the resource partition during serving. Integrated with the designs mentioned above, \nickname achieves much lower latency than the SOTA systems, \hk{reducing the average end-to-end latency per request by 1.27-2.58$\times$ on DeepSeek series models, and serves an average of 1.55-1.72$\times$ more requests under the given SLOs on Llama series models.}


\bibliographystyle{unsrt}  
\bibliography{templateArxiv}

\end{document}